\documentclass{article} 
\usepackage{iclr2025_conference,times}

\usepackage{amsmath,amsfonts,bm}

\def\eqref#1{equation~\ref{#1}}

\def\1{\bm{1}}

\DeclareMathAlphabet{\mathsfit}{\encodingdefault}{\sfdefault}{m}{sl}
\SetMathAlphabet{\mathsfit}{bold}{\encodingdefault}{\sfdefault}{bx}{n}

\usepackage{hyperref}
\usepackage{graphicx}
\usepackage{enumitem}
\usepackage{booktabs}
\usepackage{multirow}
\usepackage{array}
\usepackage{url}
\usepackage{makecell}

\usepackage{listings}

\usepackage{algorithm}
\usepackage{algpseudocode}
\usepackage{amsmath}
\usepackage{adjustbox}

\newcommand{\multitag}{\textbf{Multi-TAG }}
\newcommand{\multitagg}{\textbf{Multi-TAG}}
\definecolor{darkgreen}{rgb}{0,0.7,0}
\definecolor{darkred}{rgb}{0.7,0,0}

\title{A Toolbox, Not a Hammer --- \textsc{Multi-TAG}: Scaling Math Reasoning with Multi-Tool Aggregation}

\author{Bohan Yao\textsuperscript{1,2} \ \ Vikas Yadav\textsuperscript{1} \\
\textsuperscript{1}ServiceNow AI \ \ \textsuperscript{2}University of Washington\\
\href{mailto:s1104@cs.washington.edu}{\texttt{s1104@cs.washington.edu}}
}

\iclrfinalcopy 
\begin{document}

\maketitle

\begin{abstract}
Augmenting large language models (LLMs) with external tools is a promising avenue for developing high-performance mathematical reasoning systems. Prior tool-augmented approaches typically finetune an LLM to select and invoke a \textit{single} tool at each reasoning step and show promising results on simpler math reasoning benchmarks such as GSM8K. However, these approaches struggle with more complex math problems that require precise reasoning over multiple steps. To address this limitation, in this work, we propose \multitagg, a {\bf Multi-T}ool {\bf AG}gregation-based framework. Instead of relying on a single tool, \multitag guides an LLM to concurrently invoke multiple tools at each reasoning step. It then aggregates their diverse outputs to verify and refine the reasoning process, enhancing solution robustness and accuracy. Notably, \multitag is a finetuning-free, inference-only framework, making it readily applicable to any LLM backbone, including large open-weight models which are computationally expensive to finetune and proprietary frontier models which cannot be finetuned with custom recipes. We evaluate \multitag on four challenging benchmarks: MATH500, AIME, AMC, and OlympiadBench. Across both open-weight and closed-source LLM backbones, \multitag consistently and substantially outperforms state-of-the-art baselines, achieving average improvements of 6.0\% to 7.5\% over state-of-the-art baselines. \footnote{\multitag GitHub will be open-sourced soon.}
\end{abstract}

\section{Introduction}

Large Language Models (LLMs) have demonstrated remarkable capabilities across a wide variety of tasks, with reasoning emerging as a core area of research \citep{jiang2023mistral, openai2023gpt4, openai2022chatgpt, yang2024qwen25math}. In particular, imbuing LLMs with the ability to perform complex mathematical reasoning remains an active challenge \citep{ahn2024large}. To address this challenge, tool-augmented LLM (TALM) frameworks such as Program-Aided Language Models (PAL) \citep{gao2023pal}, Program-of-Thought (PoT) \citep{chen2022program}, Tool-Integrated Reasoning Agent (ToRA) \citep{goutora}, and MATHSENSEI \citep{das2024mathsensei} equip LLMs with external tools such as Python code execution or WolframAlpha querying. While previous TALM frameworks have shown notable progress on simpler math benchmarks such as GSM8K \cite{cobbe2021gsm8k}, their performance plateaus and remains low on more complex benchmarks such as MATH500 \citep{lightman2023lets}, AIME, AMC, and OlympiadBench \citep{he2024olympiadbench}. 

Inference-time compute scaling approaches such as OpenAI o1 \citep{OpenAI2024o1}, DeepSeek-R1 \citep{deepseek2025r1} and rStar-Math \citep{guan2025rstarmath} allocate more computational resources to LLMs at inference time to allow them to reason more methodically about problems, and also demonstrate promise towards addressing the challenge of LLM math reasoning. Although these approaches achieve impressive performance on complex math benchmarks, they generally require extensive finetuning, which can be especially brittle and requires training tricks and carefully tuned hyperparameters to be effective \citep{zeng2025simplerl}. Moreover, finetuning can be prohibitively expensive, especially for larger LLMs, and may not be possible for proprietary LLMs whose APIs do not support the finetuning recipes used in these approaches. Furthermore, recent work has shown that these approaches exhibit non-robust reasoning behaviors in various scenarios, such as number transformations \citep{yu2025benchmarking}, adversarial triggers \citep{rajeev2025cats}, unanswerable tasks \citep{hashemi2025dna}, and executing algorithmic solutions \citep{shojaee2025illusionofthinking}.

In this work, we propose \multitagg, a \textbf{Multi-T}ool \textbf{AG}gregation framework that aims to address the limitations of existing TALM frameworks on complex math reasoning tasks by adopting the inference-time scaling paradigm. As opposed to previous TALM frameworks which select a single tool to invoke at each reasoning step, \multitag scales up inference-time compute usage by invoking multiple tools at each reasoning step and aggregating their outputs, utilizing the consensus between different tool invocations to ensure that accurate reasoning steps are made.

The core benefit of multi-tool aggregation is cross-validation of different tools' outputs. Since different tools have different strengths and failure modes (see Section \ref{sec:problem_subject_analyses}), their agreement on a result provides strong evidence of its correctness. For example, a natural language reasoning tool and a Python code execution tool both proposing a consistent result lends significant credence to the validity of the result, as it is improbable that both tools made unique mistakes thematic to their own weaknesses (e.g. calculation mistakes for the natural language tool and logical mistakes for the Python tool) yet coincidentally arrived at the same incorrect result. By invoking and aggregating a diverse set of tools at each reasoning step, \multitag harnesses this principle to self-validate candidates for each step, significantly improving the overall reasoning performance.

Beyond superior reasoning performance, another key strength of \multitag is that it is a purely inference-time approach, making it readily applicable to any general instruction-tuned LLM. In contrast, finetuning-based TALM approaches such as ToRA \citep{goutora} or MathCoder \citep{wang2024mathcoder} incur a significant compute overhead for finetuning when applied to new backbone LLMs and cannot be applied to proprietary LLMs lacking finetuning APIs. To demonstrate \multitagg's transferability to different LLM backbones, we replicate our main results on three LLMs, including both open-weight and proprietary models, and observe consistent and significant performance improvements over baselines for all three LLMs. Furthermore, \multitagg's inference compute costs are adjustable, allowing users to tune \multitag to fit their cost-performance tradeoff requirements, as opposed to prior TALM approaches where compute allocation is not easily tunable. We study how performance and compute usage vary for different \multitag hyperparameter settings in Section \ref{sec:hyperparameters} and propose a simple hyperparameter heuristic to maximize performance under a fixed compute budget.

Our key contributions are:
\begin{enumerate}[noitemsep,nolistsep,left=0pt]
\item We introduce \multitagg, a tool-augmented LLM framework that effectively solves complex math reasoning tasks by aggregating multiple tool invocations at each reasoning step. The code repository for \multitag code will be open-sourced in the near future.

\item We present extensive evaluations of \multitagg, seven simple baselines, and five state-of-the-art TALM baselines on three LLM backbones and four challenging math reasoning benchmarks: MATH500, AIME, AMC, and OlympiadBench. Across the three LLMs, state-of-the-art TALM baselines perform poorly, with the strongest TALM baseline for each model achieving 1.3\% to 6.2\% \textit{lower} accuracy compared to the strongest simple baseline for each model. Meanwhile, \multitag performs much better, achieving 6.0\% to 7.5\% higher accuracy compared to the strongest simple baseline for each model and 7.9\% to 13.7\% higher accuracy compared to the strongest TALM baseline for each model.

\item We present comprehensive analyses highlighting the strengths of \multitag and benchmarking the cost-performance trade-off. In particular, we analyze \multitagg's performance by problem difficulty and problem subject area in Section \ref{sec:improvement_areas}. We also study how \multitagg's performance and compute cost vary with different hyperparameter settings and present insights for how users can tune \multitag to extract maximal performance under various computational budgets in Section \ref{sec:hyperparameters}. 
\end{enumerate}
\section{Related Work}

\paragraph{Tool-Augmented Language Models}

Recent advancements in developing tool-augmented language models (TALMs) have shown promise in improving performance on reasoning tasks. Frameworks such as ToolFormer \citep{schick2023toolformer}, OlaGPT \citep{xie2023olagpt}, and ART \citep{paranjape2023art} demonstrate that allowing LLMs to access external tools can significantly boost performance in various domains. Many previous works have also focused on developing TALMs for math reasoning tasks. Program-aided Language Models (PAL) \citep{gao2023pal} and Program of Thoughts (PoT) \citep{chen2022program} propose to prompt LLMs to generate Python code solutions to math problems instead of natural language Chain-of-Thought (CoT) solutions and achieve strong performance on simple math word problem datasets such as GSM8K \citep{cobbe2021gsm8k} and SVAMP \citep{patel-etal-2021-nlp}. Mammoth \citep{yue2023mammoth} finetunes models to produce both CoT solutions and Python code solutions to math problems. More recent works have recognized the importance of multi-step tool use for solving more challenging math problems which are too complex to solve with a single tool call. MuMath-Code \citep{yin2024mumath} and ToRA \citep{goutora} finetune models to generate solutions with multiple interleaved CoT and Python code reasoning traces. MathSensei \citep{das2024mathsensei} prompts LLMs to solve math problems by following handcrafted workflows of sequences of tool invocations. Code-based Self-Verification prompting \citep{zhou2023csv} leverages GPT4 Code Interpreter's built-in code execution capabilities to verify each CoT reasoning step with a Python script. 

While previous TALM approaches have been effective on simpler math reasoning tasks, their use of only one tool invocation per reasoning step may limit their scalability to more complex math problems. To address this limitation, we hypothesize that allowing TALMs to invoke multiple tools at each reasoning step improves their capacity for solving complex math problems by enabling cross-verification and leveraging the complementary strengths of different tools. To this end, the proposed \multitag framework aggregates outputs from multiple tools at each step, allowing different tools to validate and reinforce each part of the reasoning process, thus enhancing the accuracy of intermediate reasoning and the final solution. Furthermore, TALM approaches that rely on finetuning are not easily generalizable to all models, requiring significant computational resources to finetune larger models \citep{deepseek2025r1}, and impossible to apply to proprietary models lacking finetuning APIs. In contrast, our framework relies solely on prompting and the inherent instruction following capabilities of LLMs, allowing it to be seamlessly applied to any LLM.

\paragraph{Inference-time Compute Scaling}

Inference-time compute scaling approaches aim to improve LLM reasoning performance by utilizing more compute at inference time. Self-consistency \citep{wang2022self} proposes a simple way to implement inference-time scaling, where multiple solutions are sampled from an LLM and the most common answer is chosen. Building on top of these repeated sampling approaches, other works have further proposed to train verifiers to better select the best solution from the sampled solutions. Some works propose to train verifiers to judge entire solutions \citep{cobbe2021trainingverifierssolvemath}, whereas others propose to train verifiers to judge individual steps of solutions \citep{lightman2023lets, wang2024mathshepherd}. Inline with step-wise verification approaches, our \multitag approach also performs verification at each reasoning step, but utilizes cross-verification between different tools' outputs instead of trained verifiers.

Inference scaling via long CoT reasoning has also been shown to be an effective technique \citep{muennighoffs1, deepseek2025r1}, although solving reasoning tasks with only CoT reasoning has limitations in various settings \citep{sprague2025cot, yu2025benchmarking, rajeev2025cats, hashemi2025dna, shojaee2025illusionofthinking}. In a complementary direction, our work explores inference-time scaling through \textit{multi-tool} aggregation to enhance the performance of TALMs. 
\section{\multitag}

\begin{figure*}[!htb]
    \centering
    \includegraphics[width=\textwidth]{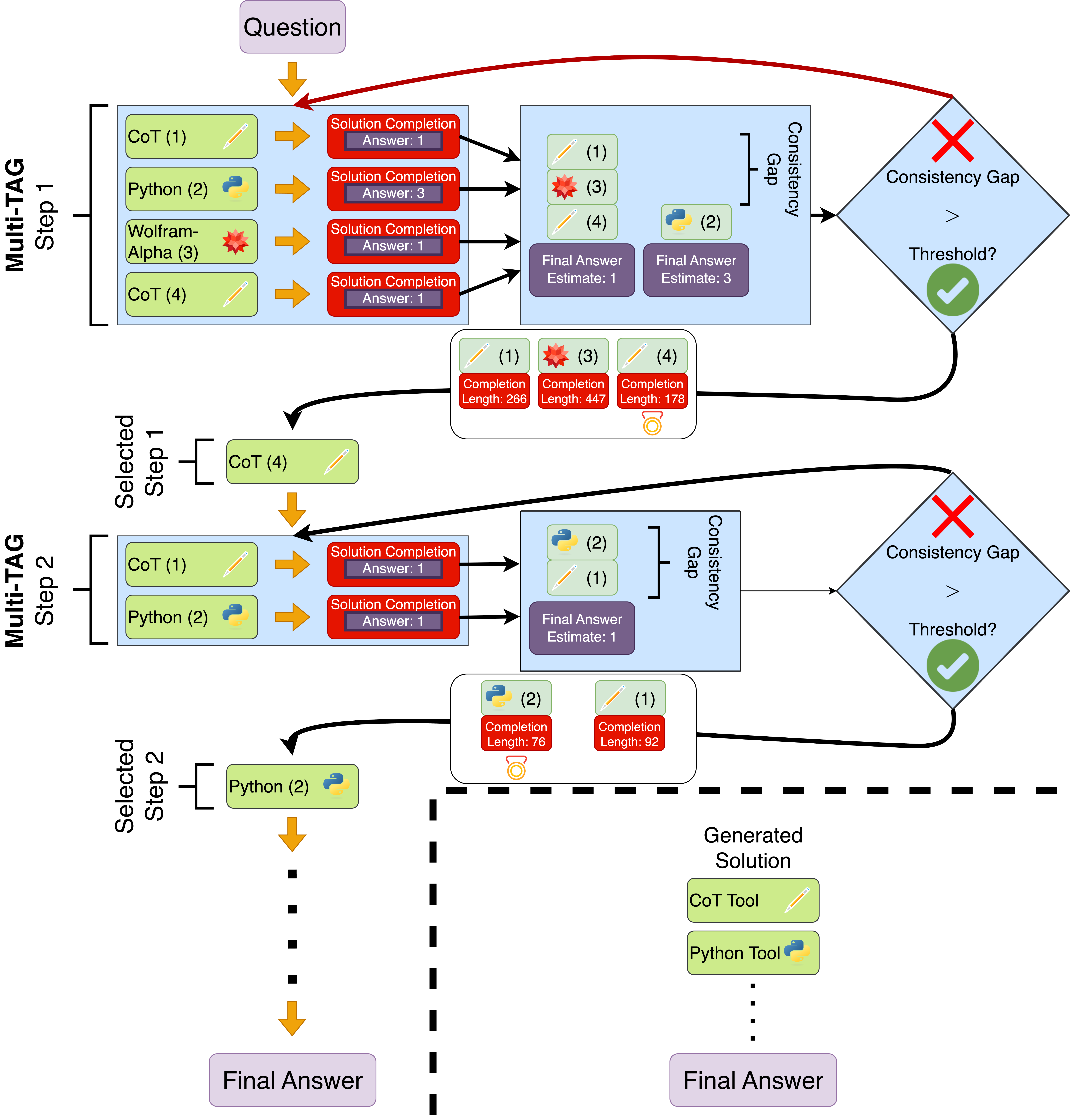}
    \caption{\footnotesize Visualization of the \multitag framework with a consistency threshold value of 1. In the first step, after the first four executors are invoked, candidates CoT (1), Python (2), WolframAlpha (3), and CoT (4) are produced. Candidates (1), (3), (4) have final answer estimate 1, while executor (2) has final answer estimate 3. The frequency of the most frequent final answer estimate, 1, is 3, while the frequency of the second most frequent final answer estimate, 3, is 1, so the consistency gap is $3-1=2$, which is greater than the consistency threshold value. Hence, executor invocation terminates. To select a candidate, first the candidates (1), (3), (4) are shortlisted as they reach the most frequent final answer estimate of 1. Then, (4) is selected as it has the shortest solution completion. In the second step, only two executors were invoked for the consistency gap to exceed the consistency threshold value. Candidate (2) was selected due to having the shorter solution completion. This process repeats until the selected step reaches a final answer for the problem. The full generated solution to the problem is the concatenation of all the selected steps.}
    \label{fig:multitool_aggregation}
\end{figure*}

\multitag is an LLM reasoning framework that scales up inference-time compute via multi-tool aggregation to improve reasoning performance. It equips an LLM-based problem solver with a diverse set of tools to solve complex problems in a step-by-step manner. At each step, each tool is invoked multiple times, generating a set of tool-augmented reasoning steps that serve as candidates for the next reasoning step. Note that each candidate may propose a different objective for the next step, creating a more diverse pool of candidates and allowing each candidate to tailor its proposed objective to the strengths of the tool used. These candidates are then aggregated, and the most accurate and productive candidate is selected to continue the solution. By aggregating candidates using different tools, \multitag exploits the unique strengths of each tool to cross-validate each candidate's reasoning. Specifically, \multitag utilizes final answer estimates derived from each candidate to aggregate candidates that utilize different tools and potentially achieve different objectives. 

Figure \ref{fig:multitool_aggregation} provides an overview of the \multitag system, and Algorithm \ref{alg:multitag} provides an explicit pseudocode implementation. Given a problem $\mathcal{P}$ and a set of tools $\mathcal{T} = \{T_1, T_2, \dots, T_t\}$, \multitag constructs a step-by-step solution $s_1, s_2, \dots, s_n$ with each step invoking one of the tools in $\mathcal{T}$. At the $p$'th step, \multitag starts by sequentially invoking a set of $m \times t$ LLM \textit{executors}. An early stopping criteria is checked after each \textit{executor's} invocation to determine if executor invocation should be terminated early; see Section \ref{sec:consistency_threshold_methods} for details. The $i$'th \textit{executor} is assigned tool $T_{((i - 1) \bmod t) + 1}$ and given $\mathcal{P}$ and the current partial solution $s_1, s_2, \dots, s_{p-1}$. It is prompted to propose a candidate $s_p^i$ for the next reasoning step. The value of $m \times t$ is a tunable hyperparameter which we call the max executors value, which can be tuned to adjust the amount of inference compute utilized.

After executor invocation is completed, each candidate $s_p^i$ is appended to the current partial solution, forming a candidate partial solution $\mathrm{cand}_{i} = s_1, s_2, \dots, s_{p-1}, s_p^i$. An LLM \textit{completer} is then invoked for each candidate partial solution. The $i$'th \textit{completer} is given $\mathcal{P}$ and $\mathrm{cand}_{i}$ and is prompted to generate a natural language solution completion $\mathrm{comp}_{i}$, which when concatenated after $\mathrm{cand}_{i}$ forms a complete solution to $\mathcal{P}$. The final answer reached by this concatenated solution serves as a quick ``approximation" of the final answer assuming $\mathrm{cand}_{i}$ is accurate, and we call it the $i$'th final answer estimate $\mathrm{est}_{i}$.

Finally, to select the best $\mathrm{cand}_{i}$ to serve as the next step in the current partial solution, a two-step selection procedure is employed. In the first step, the most frequent final answer estimate $\mathrm{maxest} = \mathrm{mode}(\{\mathrm{est}_{1}, \mathrm{est}_{2}, \dots, \mathrm{est}_{m \times t}\})$ is identified, and all candidates $\mathrm{cand}_{i}$ such that $\mathrm{est}_{i} = \mathrm{maxest}$ are shortlisted. Similar to self consistency's motivation, the more candidates that reach consistent final answers, the more confident we can be about the candidates' accuracy. In the second step, the shortlisted candidate with the shortest solution completion (measured in number of LLM tokens) is selected to be the next step in the current partial solution. Intuitively, selecting this step would lead to the most concise solution, improving \multitagg's compute efficiency. Furthermore, we find in our ablation analyses in Section \ref{sec:next_step_selection} that using this second selection step also improves performance. This finding can be intuitively explained by the Occam's Razor principle, that concise explanations should be favored over complex ones, and is consistent with recent work demonstrating that concise reasoning can sometimes lead to stronger performance \citep{sui2025stop}.

We perform ablation tests on our two-step selection procedure in Section \ref{sec:next_step_selection} and find empirically that both steps are necessary to achieve maximum performance. Furthermore, we demonstrate that the second step additionally improves inference efficiency, and removing it results in substantially higher inference costs.

\subsection{Consistency Threshold And Early Termination}
\label{sec:consistency_threshold_methods}

At each reasoning step, \multitag invokes executors sequentially. After each executor invocation, \multitag uses the consistency threshold to determine whether executor invocation should be terminated early. 

We define the consistency gap as the difference between the frequencies of the most frequent and second most frequent final answer estimates. If the consistency gap exceeds the consistency threshold value, executor invocation is terminated. Intuitively, when the consistency gap is high, the executors are largely consistent with each other, and hence we can be confident that the largest group of consistent executors are accurate. The consistency threshold is a hyperparameter that can be tuned to trade inference cost for performance and vice versa.

We perform ablation tests to validate the effectiveness of the consistency threshold in Section \ref{sec:consistency_threshold}. We find that the consistency threshold lowers inference costs substantially, while incurring a negligible accuracy penalty.

\begin{algorithm}
\caption{Pseudocode for \multitag algorithm}\label{alg:multitag}
\begin{algorithmic}[1]
\Require Problem $\mathcal{P}$, Toolset $\mathcal{T} = \{T_1, T_2, \dots, T_t\}$, Max executors value $m \times t$, Consistency threshold value $\texttt{thresh}$
\Ensure Step-by-step solution $\mathcal{S}_n = [s_1, s_2, \dots, s_n]$ to $\mathcal{P}$

\State Initialize current partial solution $\mathcal{S}_0 \gets []$
\For{$p = 0$ to $n-1$}
    \State Initialize candidate pool $\mathcal{C} \gets []$
    
    \For{$k = 1$ to $m \times t$}
        \State Invoke $k$'th \textit{executor} to generate candidate $c_{p+1}^k$ using tool $T_{((k - 1) \bmod t) + 1}$, given $\mathcal{P}$ and $\mathcal{S}_p$
        \State Append $c_{p+1}^k$ to $\mathcal{C}$
        \State Form candidate partial solution $\mathrm{cand}_{k} \gets [\mathcal{S}_p, c_{p+1}^k]$
        \State Use \textit{completer} to generate natural language solution completion $\mathrm{comp}_k$ given $\mathcal{P}$ and $\mathrm{cand}_{k}$
        \State Extract final answer estimate $\mathrm{est}_k$ from $\mathrm{comp}_k$
        \If{Consistency gap $ > \texttt{thresh}$}
            \State \textbf{break}
        \EndIf
    \EndFor
    \State Identify most frequent final answer estimate $\mathrm{maxest}$ among all $\mathrm{est}_i$
    \State Shortlist candidates $\mathcal{C}_{\mathrm{shortlist}} = \{c_{p+1}^k \mid \mathrm{est}_k = \mathrm{maxest}\}$
    \State Select $c_{p+1}^*$ from $\mathcal{C}_{\mathrm{shortlist}}$ whose $\textrm{comp}_*$ is shortest
    \State Append $c_{p+1}^*$ to current partial solution: $\mathcal{S}_{p+1} \gets [\mathcal{S}_{p}, c_{p+1}^*]$
\EndFor

\State \Return $\mathcal{S}_n = [s_1, s_2, \dots, s_n]$

\end{algorithmic}
\end{algorithm}
 
\section{Results}

\subsection{Datasets \& Models}

Following many recent works on LLM reasoning, we evaluate \multitag on challenging short answer math problems. One motivation for this choice is that there are many math datasets with thoroughly vetted ground truth answers publicly available. Furthermore, answers to short answer math problems can be easily and accurately verified by comparing the model answers to ground truth answers with a symbolic equality checker such as SymPy. Hence, evaluating on short answer math problems enhances the reliability and reproducibility of our evaluations.

We select four challenging math reasoning datasets for evaluation: MATH \cite{hendrycksmath2021},
AMC \footnote{\url{https://huggingface.co/datasets/AI-MO/aimo-validation-amc}}, AIME \footnote{\url{https://huggingface.co/datasets/AI-MO/aimo-validation-aime}}, and OlympiadBench \cite{he2024olympiadbench}. Following recent work on math reasoning, we evaluate on the MATH500 subset \cite{lightman2023lets} of the full MATH dataset to accelerate evaluations. For OlympiadBench, we only use the two English, text only, open ended (short answer) splits, OE\_TO\_maths\_en\_COMP and OE\_TO\_physics\_en\_COMP.

\multitag is applicable to any LLM. We run evaluations using LLaMA-3-70B \cite{team2024llama}, LLaMA-3.3-70B, and GPT-4o (05-13) to illustrate the efficacy of our method on less performant open models, near-frontier level open models, and frontier level proprietary models.

\subsection{\multitag Implementation Details}

For all \multitag experiments, unless specified otherwise, we use a maximum of 12 executors per step and a consistency threshold of 2. We selected these hyperparameters to achieve near-optimal performance while maintaining a relatively low compute cost; see Section \ref{sec:hyperparameters} for a detailed analysis of \multitag hyperparameters. We provide \multitag with three tools: CoT reasoning, Python script execution, and WolframAlpha queries. For sampling tool invocations, we use temperature 0.7 and top\_p 0.9, and for partial solution completions, we use temperature 0.0. For MATH500, AIME, and AMC, we use Math-Verify to grade model predictions. For OlympiadBench, we use the autograder provided in the OlympiadBench GitHub. \multitag prompts are available in Appendix \ref{sec:appendix_a}.

\subsection{Baselines}

\paragraph{Simple} We evaluate simple baselines that perform a single tool invocation per problem. Specifically, for each of the three tools available to \multitagg, we create a baseline where an LLM is prompted to solve the problem with a single invocation of the tool, with the tool's output serving as the final answer. For the WolframAlpha baseline, the WolframAlpha API often returns improperly formatted results, such as Unicode math expressions, which are erroneously marked incorrect by the autograder. To address this, we add a second LLM step to the WolframAlpha baseline to reformat the WolframAlpha output into a \LaTeX~ formatted answer. In addition, we evaluate four simple majority voting baselines where multiple single-tool solutions are sampled, and the most frequent tool output is taken as the model's answer. We create a majority voting baseline for each of the three tools available to \multitagg, each of which only sample solutions using the tool it corresponds to. A fourth majority voting baseline is developed that samples solutions using all three tools. For all majority voting baselines, we sample 12 traces per problem to match the maximum of 12 executors per step used in \multitagg. The multi-tool majority voting baseline achieves this by sampling four traces from each of the three tools.

\paragraph{Tool Augmented Frameworks} We also compare \multitag against several state-of-the-art tool augmented LLM frameworks, including PAL \citep{gao2023pal}, PoT \citep{chen2022program}, ToRA \citep{goutora}, MATHSENSEI \citep{das2024mathsensei}, and ReAct \citep{yao2022react}. Since the original ToRA work only finetuned older models, such as LLaMA-2 or CodeLLaMA, which are not used in our study, we adapt its approach by prompting newer models to emulate the ToRA reasoning process, using the few-shot prompt from the ToRA paper (which was originally used to generate ToRA traces for the training data). For MATHSENSEI, we use the PG+WA+SG setting, which was reported to achieve the highest accuracy on MATH in the original work. For ReAct, we provide the same three tools available to \multitag and write a custom prompt for using these tools.

For all non-majority voting baselines, LLM generations are conducted at temperature 0.0. For the majority voting baselines, LLM generations are conducted at temperature 0.7 and top\_p 0.9. The prompts used for all simple baselines are provided in Appendix \ref{sec:appendix_a}. Prompts for the TALM baselines are available in the \multitag GitHub \footnote{\footnotesize Will be released soon}

\subsection{Main Results}

Table \ref{tab:main_results} presents the results of the baselines and \multitag and demonstrates the superior performance of \multitag at solving challenging math reasoning problems. Over the three LLMs, the TALM baselines consistently underperform even the simple baselines, demonstrating the inability of these frameworks to address complex math problems. In contrast, \multitag outperforms all baselines on all four benchmarks and all three LLMs, demonstrating the effectiveness of multi-tool aggregation at improving the math reasoning abilities of LLMs. When compared to the strongest baseline for each LLM, \multitag achieves an average accuracy improvement of 6.6\% with LLaMA-3-70B, 6.0\% with LLaMA-3.3-70B, and 7.5\% with GPT-4o. The improvements are even more substantial when comparing only to the strongest TALM baselines, with improvements rising to 7.9\%, 8.4\%, and 13.7\%, respectively. Furthermore, the consistent improvements achieved by \multitag over both open-weight (LLaMA) and proprietary (GPT-4o) models demonstrate its generalizability to different LLM backbones.

\begin{table*}[!ht]
\centering
\resizebox{\textwidth}{!}{%
\begin{tabular}{>{\raggedright\arraybackslash}m{1.1cm} l c c c c c}
\toprule
\textbf{Model} & \textbf{Method} & \textbf{MATH500} & \textbf{AIME} & \textbf{AMC} & \textbf{OlympiadBench} & \textbf{Average} \\
\midrule
\multirow{12}{*}{\rotatebox[origin=c]{90}{\Large LLaMA-3-70B}} 
& CoT & 52.2\% & 1.1\% & 26.5\% & 16.6\% & 24.1\% \\
& Python & 45.2\% & 7.7\% & 27.7\% & 17.9\% & 24.6\% \\
& WolframAlpha Query & 23.4\% & 0.0\% & 8.4\% & 6.4\% & 9.6\% \\
& CoT MV & 58.8\% & 2.2\% & 27.7\% & 21.1\% & 27.5\% \\
& Python MV & 52.0\% & 10.0\% & 26.5\% & 21.4\% & 27.5\% \\
& WolframAlpha MV & 25.2\% & 0.0\% & 12.0\% & 7.3\% & 11.1\% \\
& CoT + Python + WolframAlpha MV & \underline{60.6\%} & 5.6\% & 33.7\% & \underline{23.6\%} & \underline{30.9\%} \\
\cmidrule(lr){2-7}
& PAL & 51.2\% & \underline{12.2\%} & \underline{36.1\%} & 18.7\% & 29.6\% \\
& PoT & 46.8\% & 8.9\% & 27.7\% & 18.6\% & 25.5\% \\
& ToRA & 54.0\% & 4.4\% & 27.7\% & 21.2\% & 26.8\% \\
& MATHSENSEI & 56.4\% & 3.3\% & 20.5\% & 14.5\% & 23.7\% \\
& ReAct & 39.4\% & 1.1\% & 13.3\% & 10.4\% & 16.1\% \\
\cmidrule(lr){2-7}
& \multitag \textbf{(Ours)} & \textbf{68.6\%} & \textbf{13.3\%} & \textbf{39.8\%} & \textbf{28.1\%} & \textbf{37.5\%} \\
\midrule
\multirow{12}{*}{\rotatebox[origin=c]{90}{\Large LLaMA-3.3-70B}}
& CoT & 75.8\% & 26.7\% & 47.0\% & 32.4\% & 45.5\% \\
& Python & 67.0\% & 28.9\% & 47.0\% & 30.0\% & 43.2\% \\
& WolframAlpha & 45.4\% & 18.9\% & 21.7\% & 12.6\% & 24.7\% \\
& CoT MV & \underline{79.0\%} & 28.9\% & 55.4\% & 36.6\% & 50.0\% \\
& Python MV & 73.0\% & \underline{35.6\%} & \underline{60.2\%} & 32.6\% & 50.4\% \\
& WolframAlpha MV & 45.6\% & 20.0\% & 22.9\% & 13.2\% & 25.4\% \\
& CoT + Python + WolframAlpha MV & \underline{79.0\%} & 33.3\% & \underline{60.2\%} & 37.6\% & \underline{52.5\%} \\
\cmidrule(lr){2-7}
& PAL & 65.8\% & 24.4\% & 47.0\% & 27.5\% & 41.2\% \\
& PoT & 70.2\% & 28.9\% & 48.2\% & 29.8\% & 44.3\% \\
& ToRA & 77.2\% & 30.0\% & 54.2\% & \underline{38.8\%} & 50.1\% \\
& MATHSENSEI & 67.4\% & 15.6\% & 30.1\% & 24.8\% & 34.5\% \\
& ReAct & 72.8\% & 17.8\% & 21.7\% & 35.4\% & 36.9\% \\
\cmidrule(lr){2-7}
& \multitag \textbf{(Ours)} & \textbf{84.2\%} & \textbf{38.9\%} & \textbf{67.5\%} & \textbf{43.5\%} & \textbf{58.5\%} \\
\midrule
\multirow{12}{*}{\rotatebox[origin=c]{90}{\Large GPT-4o}} 
& CoT & 79.6\% & 10.0\% & 47.0\% & 32.5\% & 42.3\% \\
& Python & 66.2\% & 22.2\% & 50.6\% & 30.2\% & 42.3\% \\
& WolframAlpha Query & 54.4\% & 4.4\% & 22.9\% & 16.6\% & 24.6\% \\
& CoT MV & 81.8\% & 12.2\% & 49.4\% & 36.5\% & 45.0\% \\
& Python MV & 74.2\% & \underline{28.9\%} & 59.0\% & 34.8\% & 49.2\% \\
& WolframAlpha MV & 56.2\% & 5.6\% & 22.9\% & 16.9\% & 25.4\% \\
& CoT + Python + WolframAlpha MV & \underline{86.0\%} & 22.2\% & \underline{60.2\%} & \underline{38.2\%} & \underline{51.7\%} \\
\cmidrule(lr){2-7}
& PAL & 64.6\% & 20.0\% & 44.6\% & 28.8\% & 39.5\% \\
& PoT & 51.2\% & 15.6\% & 36.1\% & 19.3\% & 30.6\% \\
& ToRA & 73.0\% & 17.8\% & 42.2\% & 32.1\% & 41.3\% \\
& MATHSENSEI & 73.4\% & 5.6\% & 43.4\% & 28.9\% & 37.8\% \\
& ReAct & 75.2\% & \underline{28.9\%} & 45.8\% & 32.1\% & 45.5\% \\
\cmidrule(lr){2-7}
& \multitag \textbf{(Ours)} & \textbf{87.0\%} & \textbf{34.4\%} & \textbf{71.1\%} & \textbf{44.1\%} & \textbf{59.2\%} \\
\bottomrule
\end{tabular}
}
\caption{\footnotesize Main results comparing \multitag with various baselines. Best score in each category is \textbf{bolded} and second best score is \underline{underlined}. MV denotes majority voting.}
\label{tab:main_results}
\end{table*}

\section{Analysis}

\subsection{\multitag Improvement Areas}
\label{sec:improvement_areas}
To better understand where \multitag improves performance relative to baselines, we compared the performances of the different methods across different MATH500 difficulty levels (ranging from 1-5 where 5 is the hardest level) and across different MATH500 problem subjects.

\subsubsection{Problem Difficulty}

\begin{figure*}[!htb]
    \centering
    \includegraphics[width=\textwidth]{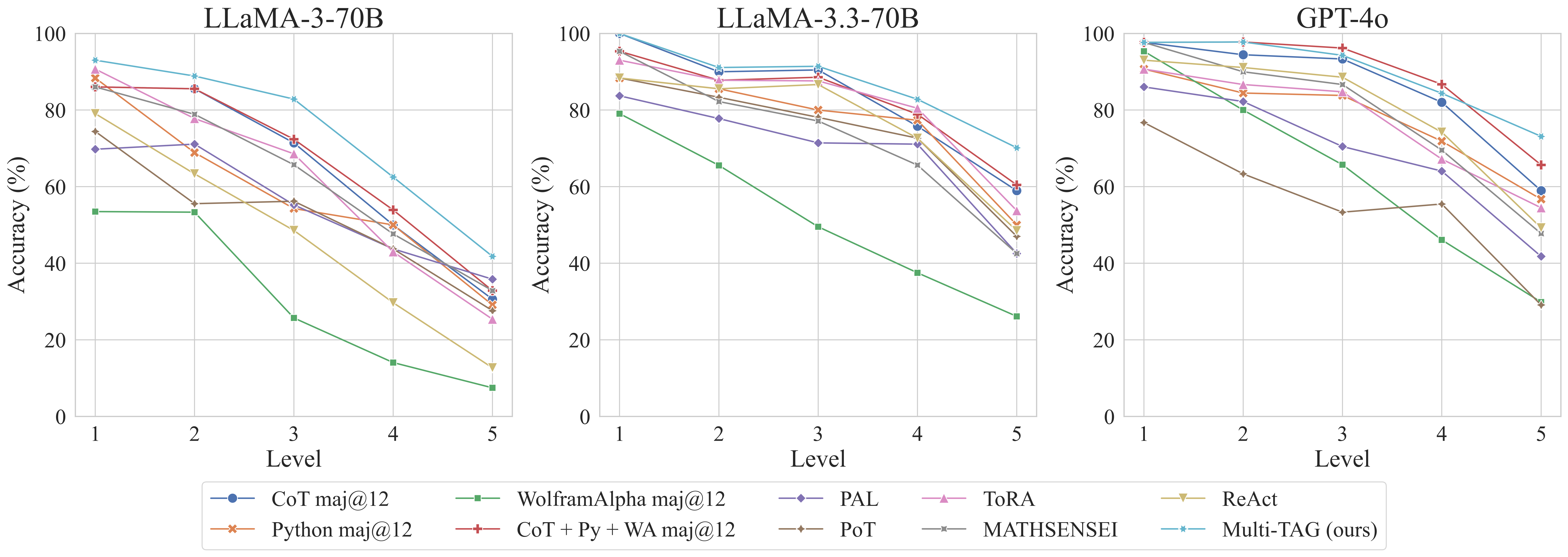}
    \caption{\footnotesize Comparison of baseline methods and \multitag on different MATH500 difficulty levels (higher levels contain more difficult problems). As shown, \multitag outperforms baselines most substantially on the more challenging problems.}
    \label{fig:acc_vs_math500lvl}
\end{figure*}

Figure \ref{fig:acc_vs_math500lvl} shows the performance of \multitag and baseline methods on different MATH500 difficulty levels. As shown, the improvements from \multitag over baselines are especially prominent at higher difficulty levels. At level 5, \multitag outperforms all baselines on LLaMA-3-70B by 6.0\%, on LLaMA-3.3-70B by 9.7\%, and on GPT-4o by 7.5\%. These improvements over previous single-tool TALM frameworks demonstrates the effectiveness of multi-tool aggregation as an inference scaling technique for boosting complex math reasoning performance.

\subsubsection{Problem Subject}
\label{sec:problem_subject_analyses}

\begin{figure*}[!htb]
    \centering
    \includegraphics[width=\textwidth]{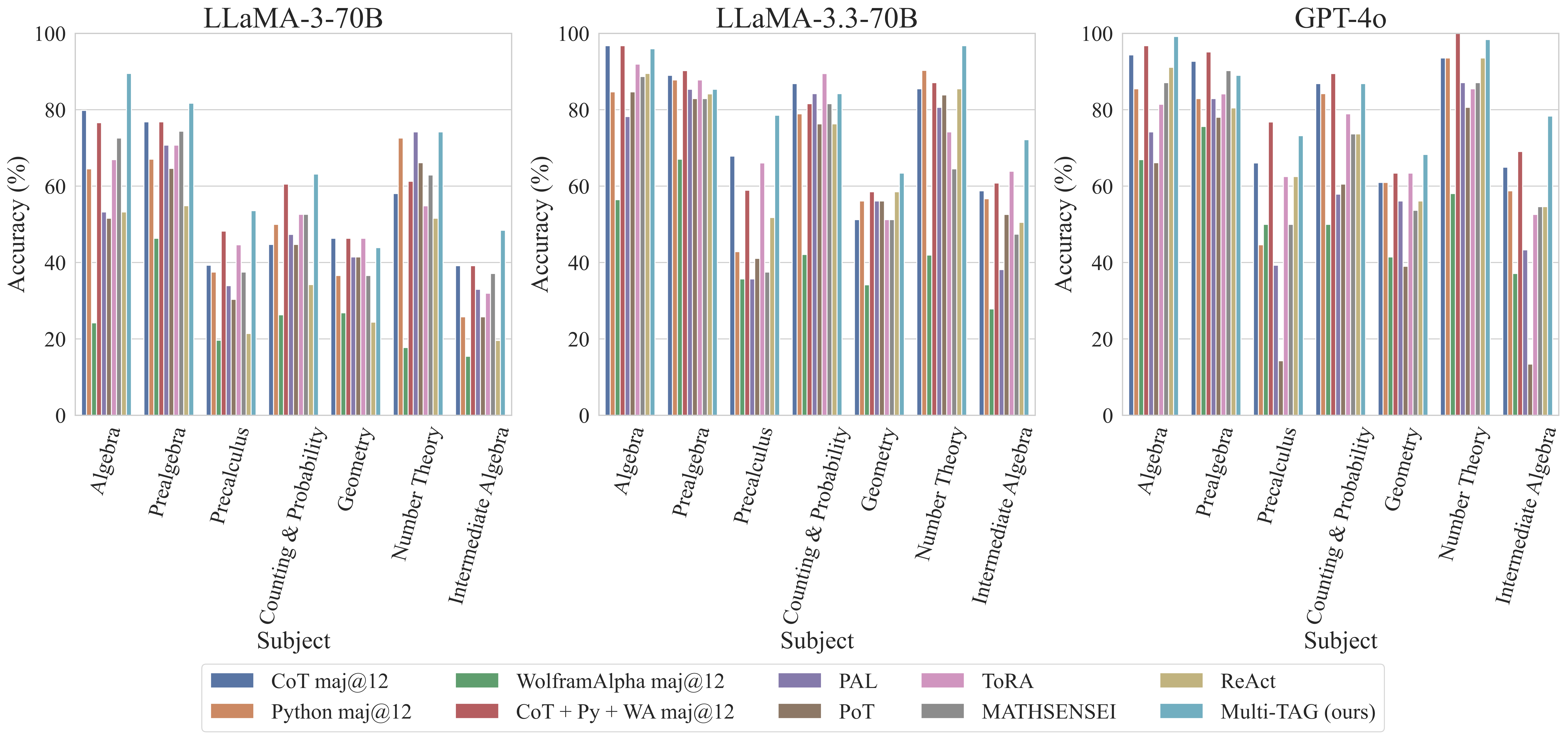}
    \caption{\footnotesize Comparison of baseline methods and \multitag on different MATH500 problem subjects. \multitag consistently performs well across subjects, outperforming all baselines on a majority of subjects.  Furthermore, simple multi-tool aggregation (CoT + Py + WA) also outperforms the three single-tool aggregation baselines on a majority of subjects.}
    \label{fig:acc_vs_subj}
\end{figure*}

Figure \ref{fig:acc_vs_subj} shows the performance of \multitag and baseline methods on different MATH500 problem subjects. \multitag outperforms all baselines in 12/21 subjects in total across the three models, demonstrating its consistent effectiveness across a diverse range of math domains. Moreover, comparing the four simple majority voting baselines, the multi-tool majority voting baseline (CoT + Py + WA) outperformed all three single-tool majority voting baselines in 12/21 subjects in total across the three models. This highlights the synergistic benefits of aggregating different tools together, improving upon the performance of aggregating each of the tools individually.

\subsection{Ablation Study}
\label{sec:ablations}

\subsubsection{Token Consumption Cost}
\label{sec:token_consumption_cost}

We verify that the improvements from \multitag over baselines are not simply a result of \multitag utilizing more LLM inference compute (i.e. using more tokens). To do so, for each of our simple majority voting baselines, we modify the number of sampled LLM traces per problem so that each baseline and \multitag have matching token consumption costs. Similarly, for each of our TALM baselines, to increase the amount of tokens used to match \multitagg, we simply sample multiple TALM traces for each problem and apply majority voting over the final answers reached by the traces.

Token consumption cost is defined as $0.25P + O$, where $P$ is the number of prompt tokens and $O$ is the number of generated output tokens. The $0.25$ weighting of prompt token cost is based on OpenAI's GPT-4o API pricing, which as of time of writing is $\$2.50$ per million prompt tokens and $\$10$ per million output tokens. 

We evaluated \multitag and our token consumption-matched baselines and report the results in Table \ref{tab:token_consumption_cost}. For cost-related reasons, we only evaluate GPT-4o as the backbone LLM for these experiments. \multitag continues to outperform the token-matched baselines, achieving superior results on all four benchmarks and achieving a 7.7\% average accuracy improvement over the strongest baseline.

\begin{table}[htbp!]
\centering
\small
\begin{tabular}{l c c c c c}
\toprule
\textbf{Method} & \textbf{MATH500} & \textbf{AIME} & \textbf{AMC} & \textbf{OlympiadBench} & \textbf{Average} \\ 
\midrule
CoT (maj@19) & \underline{84.2}\% & 10.0\% & 51.8\% & 37.7\% & 45.9\% \\
Python (maj@35) & 75.2\% & 30.0\% & \underline{63.9}\% & 35.4\% & 51.1\% \\
WolframAlpha (maj@70) & 55.6\% & 7.8\% & 22.9\% & 15.5\% & 25.5\% \\
CoT + Python + WolframAlpha (maj@33) & \underline{84.2}\% & 22.2\% & 60.2\% & 39.0\% & 51.4\% \\
\midrule
PAL (maj@34) & 71.8\% & 27.8\% & 57.8\% & 32.5\% & 47.5\% \\
PoT (maj@14) & 75.4\% & \underline{32.2}\% & 61.4\% & 35.8\% & 51.2\% \\
ToRA (maj@6) & 82.2\% & 27.8\% & 55.4\% & \underline{40.7}\% & \underline{51.5}\% \\
MATHSENSEI (maj@3) & 78.4\% & 13.3\% & 44.6\% & 29.2\% & 41.4\% \\
ReAct (maj@6) & 81.0\% & 22.2\% & 62.7\% & 38.2\% & 51.0\% \\
\midrule
\multitag \textbf{(Ours)} & \textbf{87.0}\% & \textbf{34.4}\% & \textbf{71.1}\% & \textbf{44.1}\% & \textbf{59.2}\% \\
\bottomrule
\end{tabular}
\caption{\footnotesize Results of \multitag and token consumption-matched baselines. The number of sampled traces per problem used for each token consumption-matched baseline is given as maj@x. For simple multi-tool majority voting (CoT + Python + WolframAlpha), the 33 traces are split evenly between CoT, Python, and WolframAlpha traces. Best score in each category is \textbf{bolded} and second best score is \underline{underlined}. GPT-4o is used as the LLM for these experiments.}
\label{tab:token_consumption_cost}
\end{table}

\subsubsection{Consistency Threshold}
\label{sec:consistency_threshold}

To verify that \multitagg's consistency threshold effectively reduces the token consumption cost while incurring minimal performance degradation, we compare the results of running \multitag with a consistency threshold of 2 and the results of running \multitag without a consistency threshold. We also vary the max executors parameter and the backbone LLM to ensure the effectiveness of the consistency threshold for all settings. We report the MATH500 accuracy and average token consumption cost (as defined in Section \ref{sec:token_consumption_cost}) per problem for each of the settings in Table \ref{tab:consistency_threshold}.

As shown, the accuracy degradation incurred by applying the consistency threshold is minimal, with a maximum degradation of $2.0\%$ and an average degradation of $0.1\%$. Meanwhile, the token consumption cost is significantly reduced in all settings, with relative reductions ranging from $14.7\%$ to $63.6\%$ and an average relative reduction of $43.8\%$. Thus, the consistency threshold effectively reduces the computational cost of \multitag without compromising its performance.

\begin{table}[htbp!]
\centering
\small
\resizebox{\textwidth}{!}{%
\begin{tabular}{l cc|cc|cc}
\toprule
& \multicolumn{2}{c}{\textbf{LLaMA-3-70B}} & \multicolumn{2}{c}{\textbf{LLaMA-3.3-70B}} & \multicolumn{2}{c}{\textbf{GPT-4o}} \\
\cmidrule(lr){2-3} \cmidrule(lr){4-5} \cmidrule(lr){6-7}
\textbf{Max Executors} & \textbf{With Threshold} & \multicolumn{1}{c}{\textbf{Without Threshold}} & \textbf{With Threshold} & \multicolumn{1}{c}{\textbf{Without Threshold}} & \textbf{With Threshold} & \textbf{Without Threshold}\\
\midrule
6 & \makecell[l]{67.0\% \textcolor{darkgreen}{$\uparrow$ 0.2\%} \\ (5361 \textcolor{darkgreen}{$\downarrow$ 14.7\%})} & \makecell[l]{66.8\% \\ (6286)}
  & \makecell[l]{82.0\% \textcolor{darkred}{$\downarrow$ 1.8\%} \\ (5967 \textcolor{darkgreen}{$\downarrow$ 37.6\%})} & \makecell[l]{83.8\% \\ (9559)}
  & \makecell[l]{82.6\% \textcolor{darkred}{$\downarrow 1.4\%$} \\ (6090 \textcolor{darkgreen}{$\downarrow 23.8\%$})} & \makecell[l]{84.0\% \\ (7989)}\\

9 & \makecell[l]{66.2\% \textcolor{darkred}{$\downarrow 0.4\%$} \\ (6746 \textcolor{darkgreen}{$\downarrow 25.8\%$})} & \makecell[l]{66.6\% \\ (9091)} & \makecell[l]{85.8\% \textcolor{darkgreen}{$\uparrow 1.4\%$} \\ (7766 \textcolor{darkgreen}{$\downarrow 44.0\%$})} & \makecell[l]{84.4\% \\ (13859)} & \makecell[l]{86.4\% \textcolor{darkgreen}{$\uparrow 1.4\%$} \\ (6157 \textcolor{darkgreen}{$\downarrow 40.9\%$})} & \makecell[l]{85.0\% \\ (10425)}\\

12 & \makecell[l]{68.6\% \textcolor{darkgreen}{$\uparrow$ 0.2\%} \\ (7916 \textcolor{darkgreen}{$\downarrow$ 34.7\%})} & \makecell[l]{68.4\% \\ (12124)}
   & \makecell[l]{84.2\% \textcolor{darkred}{$\downarrow$ 2.0\%} \\ (7945 \textcolor{darkgreen}{$\downarrow$ 56.3\%})} & \makecell[l]{86.2\% \\ (18190)}
   & \makecell[l]{87.0\% \textcolor{darkgreen}{$\uparrow 0.8\%$} \\ (7952 \textcolor{darkgreen}{$\downarrow 48.3\%$})} & \makecell[l]{86.2\% \\ (15376)}\\

15 & \makecell[l]{68.6\% \textcolor{darkred}{$\downarrow 1.2\%$} \\ (8891 \textcolor{darkgreen}{$\downarrow 40.4\%$})} & \makecell[l]{69.8\% \\ (14922)} & \makecell[l]{86.0\% \textcolor{darkgreen}{$\uparrow 0.6\%$} \\ (9274 \textcolor{darkgreen}{$\downarrow 58.8\%$})} & \makecell[l]{85.4\% \\ (22501)} & \makecell[l]{87.6\% \textcolor{darkgreen}{$\uparrow 1.2\%$} \\ (8214 \textcolor{darkgreen}{$\downarrow 52.0\%$})} & \makecell[l]{86.4\% \\ (17127)}\\

18 & \makecell[l]{67.8\% \textcolor{darkred}{$\downarrow$ 1.2\%} \\ (9918 \textcolor{darkgreen}{$\downarrow$ 56.7\%})} & \makecell[l]{69.0\% \\ (17507)}
   & \makecell[l]{86.6\% \textcolor{darkgreen}{$\uparrow$ 1.0\%} \\ (9727 \textcolor{darkgreen}{$\downarrow$ 63.6\%})} & \makecell[l]{85.6\% \\ (26743)}
   & \makecell[l]{85.6\% \textcolor{darkred}{$\downarrow 0.4\%$} \\ (9005 \textcolor{darkgreen}{$\downarrow 58.8\%$})} & \makecell[l]{86.0\% \\ (21883)}\\
\bottomrule
\end{tabular}
}
\caption{\footnotesize MATH500 scores and average token consumption costs (as defined in Section \ref{sec:token_consumption_cost}) per problem of \multitag with and without the consistency threshold. Token consumption costs are in (parentheses).}
\label{tab:consistency_threshold}
\end{table}

\subsubsection{Candidate Step Selection}
\label{sec:next_step_selection}

To verify the efficacy of the candidate step selection algorithm, we study the effects of simplifying the procedure on performance and token consumption cost (as defined in Section \ref{sec:token_consumption_cost}). Recall that the procedure consists of the following two steps:

\begin{enumerate}[noitemsep,nolistsep,left=0pt,label={(\textbf{\arabic*})}]
    \item Identify the most frequent candidate final answer and mark all candidate steps reaching this final answer.
    \item From the marked candidates, select the candidate with the shortest solution completion.
\end{enumerate}

We compare four approaches: Full (the unmodified algorithm from \multitag with both \textbf{(1)} and \textbf{(2)}), Answer Only (replacing \textbf{(2)} with randomly selecting a marked candidate), Length Only (replacing \textbf{(1)} with marking all candidates), and Random (select a random candidate without using either \textbf{(1)} or \textbf{(2)}). We evaluate \multitag with each of the modified candidate selection procedures on MATH500 and report the results in Table \ref{tab:next_step_selection}.

As shown, all of the simplified candidate selection procedures significantly underperform the Full procedure. On average, the performance degradation is $2.6\%$ for Answer Only, $5.5\%$ for Length Only, and $7.9\%$ for Random. This demonstrates the necessity of both steps of the algorithm to maximize performance. Furthermore, the results show the isolated contribution of \textbf{(2)} to computational efficiency. The only difference between Full and Answer Only is the inclusion of \textbf{(2)} in the former, which reduces the token consumption cost by $27.9\%$ on average. Similarly, the only difference between Length Only and Random is the inclusion of \textbf{(2)} in the former, which reduces the token consumption cost by $25.5\%$ on average. These results demonstrate that \textbf{(2)} additionally improves the computational efficiency of \multitagg.

\begin{table}[htbp!]
\centering
\small
\resizebox{\textwidth}{!}{%
\begin{tabular}{l c c c}
\toprule
\textbf{Next Step Selection Procedure} & \textbf{LLaMA-3-70B} & \textbf{LLaMA-3.3-70B} & \textbf{GPT-4o} \\ 
\midrule
Full & \makecell[c]{68.6\% \\ (7916)} & \makecell[c]{84.2\% \\ (7945)} & \makecell[c]{87.0\% \\ (7952)} \\
\midrule
Answer Only & \makecell[c]{64.8\% \textcolor{darkred}{$\downarrow 3.8\%$} \\ (9492 \textcolor{darkred}{$\uparrow 19.9\%$})} & \makecell[c]{83.2\% \textcolor{darkred}{$\downarrow 1.0\%$} \\ (11538 \textcolor{darkred}{$\uparrow 45.2\%$})} & \makecell[c]{84.0\% \textcolor{darkred}{$\downarrow 3.0\%$} \\ (9420 \textcolor{darkred}{$\uparrow 18.5\%$})} \\
Length Only & \makecell[c]{56.8\% \textcolor{darkred}{$\downarrow 11.8\%$} \\ (7548 \textcolor{darkgreen}{$\downarrow 4.6\%$})} & \makecell[c]{82.0\% \textcolor{darkred}{$\downarrow 2.2\%$} \\ (9325 \textcolor{darkred}{$\uparrow 17.4\%$})} & \makecell[c]{84.6\% \textcolor{darkred}{$\downarrow 2.4\%$} \\ (7412 \textcolor{darkgreen}{$\downarrow 6.8\%$})} \\
Random & \makecell[c]{54.8\% \textcolor{darkred}{$\downarrow 13.8\%$} \\ (10023 \textcolor{darkred}{$\uparrow 26.6\%$})} & \makecell[c]{78.8\% \textcolor{darkred}{$\downarrow 5.4\%$} \\ (12635 \textcolor{darkred}{$\uparrow 59.0\%$})} & \makecell[c]{82.4\% \textcolor{darkred}{$\downarrow 4.6\%$} \\ (9976 \textcolor{darkred}{$\uparrow 25.5\%$})} \\
\bottomrule
\end{tabular}
}
\caption{\footnotesize MATH500 scores and average token consumption cost (as defined in Section \ref{sec:token_consumption_cost}) per problem of \multitag with the proposed and simplified candidate step selection procedures. Token consumption costs are in (parentheses).}
\label{tab:next_step_selection}
\end{table}

\subsection{Hyperparameters Study}
\label{sec:hyperparameters}

We investigated the influence of \multitagg's two primary hyperparameters-the maximum number of executors and the consistency threshold value-on its performance and computational cost. We evaluate \multitag with various hyperparameter configurations and with all three backbone LLMs on MATH500. The results are reported in Table \ref{tab:hyperparameters}.

The results show a strong, statistically significant positive correlation between performance and the max executors value. The Spearman correlation coefficients were .832 ($p < .01$), .535 ($p = .04$), and .549 ($p = .03$) for LLaMA-3-70B, LLaMA-3.3-70B, and GPT-4o results, respectively. In contrast, the consistency threshold value showed no statistically significant correlation with performance, with coefficients of .057 ($p = .84$), .028 ($p = .92$), and .162 ($p = .56$). Thus, to increase performance, the max executors value should be increased.

While increasing the max executors value boosts performance, it also significantly increases computational costs. The results demonstrate the crucial role of the consistency threshold to mitigate this increase. For instance, when increasing max executors from 6 to 18, the average increase in token consumption cost across all models was only 49.3\% with a consistency threshold of 1. This cost increase was substantially higher for thresholds of 2 (65.3\%) and 3 (70.4\%). This demonstrates that lower consistency threshold values effectively contain costs, especially for larger max executors settings.

These findings suggest a simple heuristic for setting \multitag hyperparameters: the max executors value should be set as high as the compute budget allows to maximize performance, then the consistency threshold value should be set to a low value, such as 1 or 2, to minimize the token consumption cost.

\begin{table}[htbp!]
\centering
\newcolumntype{C}[1]{>{\centering\arraybackslash}p{#1}}
\begin{tabular}{C{0.05\textwidth} C{0.15\textwidth} C{0.09\textwidth} C{0.09\textwidth} C{0.09\textwidth} C{0.09\textwidth} C{0.09\textwidth}}
\toprule
\textbf{Model} & \textbf{Consistency} & \multicolumn{5}{c}{\textbf{Max Executors}} \\
\cmidrule(lr){3-7}
& \textbf{Threshold} & \textbf{6} & \textbf{9} & \textbf{12} & \textbf{15} & \textbf{18} \\
\midrule
\multirow{2}{*}{\rotatebox[origin=c]{90}{\small LLaMA-3-70B}} 
& \textbf{1} & \makecell[c]{63.0\% \\ (4518)} & \makecell[c]{67.2\% \\ (5592)} & \makecell[c]{66.2\% \\ (6565)} & \makecell[c]{67.8\% \\ (7272)} & \makecell[c]{69.2\% \\ (7799)} \\
& \textbf{2} & \makecell[c]{67.0\% \\ (5361)} & \makecell[c]{66.2\% \\ (6746)} & \makecell[c]{68.6\% \\ (7916)} & \makecell[c]{68.6\% \\ (8891)} & \makecell[c]{67.8\% \\ (9918)} \\
& \textbf{3} & \makecell[c]{63.8\% \\ (5838)} & \makecell[c]{66.6\% \\ (7380)} & \makecell[c]{67.2\% \\ (9267)} & \makecell[c]{68.0\% \\ (10074)} & \makecell[c]{68.8\% \\ (11418)} \\
\midrule
\multirow{5}{*}{\raisebox{0.15em}[0pt][0pt]{\rotatebox[origin=c]{90}{\small LLaMA-3.3-70B}}}
& \textbf{1} & \makecell[c]{84.0\% \\ (5023)} & \makecell[c]{84.8\% \\ (5913)} & \makecell[c]{83.6\% \\ (6400)} & \makecell[c]{86.2\% \\ (6752)} & \makecell[c]{85.0\% \\ (6793)} \\
& \textbf{2} & \makecell[c]{82.0\% \\ (5967)} & \makecell[c]{85.8\% \\ (7766)} & \makecell[c]{84.2\% \\ (7945)} & \makecell[c]{86.0\% \\ (9274)} & \makecell[c]{86.6\% \\ (9727)} \\
& \textbf{3} & \makecell[c]{83.8\% \\ (6460)} & \makecell[c]{86.0\% \\ (9061)} & \makecell[c]{84.6\% \\ (9289)} & \makecell[c]{85.6\% \\ (10497)} & \makecell[c]{84.4\% \\ (10555)} \\
\midrule
\multirow{4}{*}{\rotatebox[origin=c]{90}{\small GPT-4o}} 
& \textbf{1} & \makecell[c]{86.0\% \\ (4711)} & \makecell[c]{85.0\% \\ (4822)} & \makecell[c]{86.0\% \\ (6098)} & \makecell[c]{86.2\% \\ (6297)} & \makecell[c]{86.6\% \\ (6594)} \\
& \textbf{2} & \makecell[c]{82.6\% \\ (6090)} & \makecell[c]{86.4\% \\ (6157)} & \makecell[c]{87.0\% \\ (7952)} & \makecell[c]{87.6\% \\ (8214)} & \makecell[c]{86.2\% \\ (9008)} \\
& \textbf{3} & \makecell[c]{85.0\% \\ (7064)} & \makecell[c]{87.0\% \\ (7352)} & \makecell[c]{86.2\% \\ (9253)} & \makecell[c]{85.6\% \\ (9556)} & \makecell[c]{87.2\% \\ (10753)} \\
\bottomrule
\end{tabular}
\caption{\footnotesize MATH500 scores and average token consumption cost (as defined in Section \ref{sec:token_consumption_cost}) per problem of various max executor, consistency threshold configurations of \multitagg. Token consumption costs are in (parentheses).}
\label{tab:hyperparameters}
\end{table}
\section{Conclusion}

In this paper, we present \multitag, a novel tool-augmented LLM framework for math reasoning. Unlike previous TALM frameworks, \multitag scales up the inference-time compute allocated by allowing the LLM to invoke and aggregate multiple tools at each reasoning step. As a result, \multitag achieves superior results on four complex math reasoning benchmarks, outperforming the strongest baselines by 6.0\% to 7.5\% over three different backbone LLMs. Furthermore, \multitag is widely applicable, enabling the use of any general instruction-tuned LLM and enabling computational costs to be tuned according to specific cost/performance requirements. Our results demonstrate that multi-tool aggregation is a promising avenue for future work on advancing LLM math reasoning capabilities.

\newpage

\bibliography{iclr2025_conference}
\bibliographystyle{iclr2025_conference}

\newpage

\appendix

\section{LLM Prompts}
\label{sec:appendix_a}

We provide the LLM prompts used for all components of \multitag in Section \ref{sec:multitag_prompts} and the prompts used for all simple baselines in Section \ref{sec:baseline_prompts}. Prompts used for TALM baselines can be found in the \multitag GitHub \footnote{\footnotesize Will be released soon}.

\subsection{\multitag Prompts}
\label{sec:multitag_prompts}

\begin{lstlisting}[frame=single, title=CoT Executor System Prompt]
You are a math problem solving agent working on solving a problem iteratively. The problem and the current progress will be given below. The current progress consists of a sequence of steps separated by "---" which may consist of natural language reasoning, Python scripts, and WolframAlpha queries. Python script execution outputs are given at the bottom of a step within ```output ```, and WolframAlpha query results are given at the bottom of a step within ```result ```. Your task is to write the next step in the solution in the form of natural language reasoning.

If the solution is complete, you may give the final answer (NOTE: you may not give the final answer if you also write a step. Only give the final answer if the solution is complete without you writing an additional step). Express the answer using LaTeX formatting and do not include units or other unnecessary text in the answer. It's okay to leave the final answer unsimplified, for example expressed as a decimal. Do not round final answers that are decimals. Make sure to read the question carefully and answer exactly what the problem is asking for. Format the answer by enclosing the answer within <final_answer></final_answer> and putting the answer within \boxed{{}}. For example:
<final_answer>
The final answer is \boxed{{[final answer formatted using LaTeX]}}
</final_answer>
\end{lstlisting}

\begin{lstlisting}[frame=single, title=Python Executor System Prompt]
You are a math problem solving agent working on solving a problem iteratively. The problem and the current progress will be given below. The current progress consists of a sequence of steps separated by "---" which may consist of natural language reasoning, Python scripts, and WolframAlpha queries. Python script execution outputs are given at the bottom of a step within ```output ```, and WolframAlpha query results are given at the bottom of a step within ```result ```. Your task is to write the next step in the solution in the form of a Python script and a brief explanation of what your script calculates.

If the solution is complete, you may give the final answer (NOTE: you may not give the final answer if you also write a step. Only give the final answer if the solution is complete without you writing an additional step). Express the answer using LaTeX formatting and do not include units or other unnecessary text in the answer. It's okay to leave the final answer unsimplified, for example expressed as a decimal. Do not round final answers that are decimals. Make sure to read the question carefully and answer exactly what the problem is asking for. Format the answer by enclosing the answer within <final_answer></final_answer> and putting the answer within \boxed{{}}. For example:
<final_answer>
The final answer is \boxed{{[final answer formatted using LaTeX]}}
</final_answer>

To write the next step, you must follow the following format:
```python
[Python script, assigning the desired output to the `result` global variable]
```
[Brief explanation of what your script calculates]
\end{lstlisting}

\begin{lstlisting}[frame=single, title=WolframAlpha Executor System Prompt]
You are a math problem solving agent working on solving a problem iteratively. The problem and the current progress will be given below. The current progress consists of a sequence of steps separated by "---" which may consist of natural language reasoning, Python scripts, and WolframAlpha queries. Python script execution outputs are given at the bottom of a step within ```output ```, and WolframAlpha query results are given at the bottom of a step within ```result ```. Your task is to write the next step in the solution in the form of a WolframAlpha query and a brief explanation of what your query calculates.

If the solution is complete, you may give the final answer (NOTE: you may not give the final answer if you also write a step. Only give the final answer if the solution is complete without you writing an additional step). Express the answer using LaTeX formatting and do not include units or other unnecessary text in the answer. It's okay to leave the final answer unsimplified, for example expressed as a decimal. Do not round final answers that are decimals. Make sure to read the question carefully and answer exactly what the problem is asking for. Format the answer by enclosing the answer within <final_answer></final_answer> and putting the answer within \boxed{{}}. For example:
<final_answer>
The final answer is \boxed{{[final answer formatted using LaTeX]}}
</final_answer>

To write the next step, you must follow the following format:
```wolfram
[WolframAlpha query]
```
[Brief explanation of what your query calculates]
\end{lstlisting}

\begin{lstlisting}[frame=single, title=Executor User Prompt]
# Problem
{problem}

# Partial Solution
{progress}

# Final Instructions
Above are the problem and potentially incomplete solution. Note that the partial solution has already been verified for accuracy, so you should assume it is correct. Write the next step or give the final answer if the partial solution is complete. Remember that you must write a step of the specified form above (or give the final answer using the specific format above). You must write a single logical step (or give the final answer), and stop after completing a single step.
\end{lstlisting}

\begin{lstlisting}[frame=single,title=Solution Completion System Prompt]
You are a math problem solver working on completing a partial solution to a problem. The problem and partial solution will be given below. The partial solution consists of a sequence of steps separated by "---" which may consist of natural language reasoning, Python scripts, and WolframAlpha queries. Python script execution outputs are given at the bottom of a step within ```output ```, and WolframAlpha query results are given at the bottom of a step within ```result ```. Your task is to continue the partial solution to finish solving the problem. You may only use natural language reasoning in your response (you may not use Python or WolframAlpha). Enclose the final answer within \boxed{}. Express the answer using LaTeX formatting and do not include units or other unnecessary text in the answer. It's okay to leave the final answer unsimplified, for example expressed as a decimal. Do not round final answers that are decimals. Make sure to read the question carefully and answer exactly what the problem is asking for.
\end{lstlisting}

\begin{lstlisting}[frame=single, title=Solution Completion User Prompt]
# Problem
{problem}

# Partial Solution
{progress}

# Final Instructions
Above are the problem and partial solution to continue. Note that the partial solution has already been verified for accuracy, so you should assume it is correct. Continue the partial solution to finish solving the problem. You don't need to follow any specific format like the step-by-step format of the partial solution.
\end{lstlisting}

\subsection{Baseline Prompts}
\label{sec:baseline_prompts}

\begin{lstlisting}[frame=single, title=CoT User Prompt]
Solve the following problem step by step. Express the final answer using LaTeX formatting and enclose it within \boxed{{}}. Do not include units or other unnecessary text in the answer.

{problem}
\end{lstlisting}

\begin{lstlisting}[frame=single, title=Python User Prompt]
Solve the following problem by writing a single Python script. Your script should be enclosed within ```python ```. You may only write a single Python script in your response. Do not include units or other unnecessary text in the answer. Your answer should be submitted by assigning the answer to the "result" global variable.

{problem}
\end{lstlisting}

\begin{lstlisting}[frame=single, title=WolframAlpha User Prompt]
Solve the following problem by writing a single WolframAlpha query. Your WolframAlpha query must be enclosed within ```wolfram ```. You may only write a single WolframAlpha query in your response.

{problem}
\end{lstlisting}

\begin{lstlisting}[frame=single, title=Convert WolframAlpha Output to LaTeX User Prompt]
Here is the output from your query:
```result
{output}
```

Now, give the final answer to the problem. Express the final answer using LaTeX formatting and enclose it within \boxed{{}}. Do not include units or other unnecessary text in the answer. Your response must only contain the final answer and nothing else.
\end{lstlisting}
\section{Costs Analysis}
\label{sec:appendix_b}

\subsection{Main Results Token Consumption Cost}
Table \ref{tab:appendix_b} reports the average token consumption cost (as defined in Section \ref{sec:token_consumption_cost}) per problem for all results in the main results (Table \ref{tab:main_results}).

\begin{table*}[!ht]
\centering
\resizebox{\textwidth}{!}{%
\begin{tabular}{>{\raggedright\arraybackslash}m{1.1cm} l c c c c c}
\toprule
\textbf{Model} & \textbf{Method} & \textbf{MATH500} & \textbf{AIME} & \textbf{AMC} & \textbf{OlympiadBench} & \textbf{Average} \\
\midrule
\multirow{12}{*}{\rotatebox[origin=c]{90}{\Large LLaMA-3-70B}} 
& CoT & 348 & 790 & 583 & 521 & 561 \\
& Python & 97 & 153 & 128 & 135 & 128 \\
& WolframAlpha Query & 130 & 194 & 158 & 168 & 163 \\
& CoT MV & 3757 & 6220 & 5127 & 5922 & 5257 \\
& Python MV & 1160 & 1832 & 1629 & 1644 & 1566 \\
& WolframAlpha MV & 1584 & 2318 & 1939 & 2049 & 1973 \\
& CoT + Python + WolframAlpha MV & 2236 & 3928 & 2868 & 3203 & 3059 \\
\cmidrule(lr){2-7}
& PAL & 280 & 355 & 379 & 318 & 333 \\
& PoT & 566 & 621 & 610 & 613 & 603 \\
& ToRA & 1261 & 1437 & 1299 & 1420 & 1354 \\
& MATHSENSEI & 2730 & 3298 & 3049 & 3139 & 3054 \\
& ReAct & 1386 & 2242 & 1784 & 1883 & 1824 \\
\cmidrule(lr){2-7}
& \multitag \textbf{(Ours)} & 7916 & 14448 & 12125 & 12567 & 11764 \\
\midrule
\multirow{12}{*}{\rotatebox[origin=c]{90}{\Large LLaMA-3.3-70B}}
& CoT & 610 & 929 & 885 & 886 & 828 \\
& Python & 206 & 457 & 319 & 250 & 308 \\
& WolframAlpha & 148 & 231 & 184 & 197 & 190 \\
& CoT MV & 7303 & 11813 & 10877 & 10558 & 10138 \\
& Python MV & 2448 & 4326 & 3946 & 3049 & 3442 \\
& WolframAlpha MV & 1808 & 3167 & 2329 & 2330 & 2409 \\
& CoT + Python + WolframAlpha MV & 3845 & 6381 & 5572 & 5286 & 5271 \\
\cmidrule(lr){2-7}
& PAL & 386 & 677 & 585 & 475 & 531 \\
& PoT & 609 & 817 & 753 & 697 & 719 \\
& ToRA & 2149 & 3824 & 2610 & 2913 & 2874 \\
& MATHSENSEI & 3974 & 5247 & 5174 & 5099 & 4874 \\
& ReAct & 1436 & 2627 & 3181 & 2295 & 2385 \\
\cmidrule(lr){2-7}
& \multitag \textbf{(Ours)} & 7945 & 17809 & 11756 & 14186 & 12924 \\
\midrule
\multirow{12}{*}{\rotatebox[origin=c]{90}{\Large GPT-4o}} 
& CoT & 582 & 992 & 909 & 837 & 830 \\
& Python & 287 & 521 & 353 & 390 & 388 \\
& WolframAlpha Query & 129 & 234 & 233 & 197 & 198 \\
& CoT MV & 6485 & 9889 & 9306 & 8885 & 8641 \\
& Python MV & 3195 & 5375 & 5142 & 4373 & 4521 \\
& WolframAlpha MV & 1614 & 2847 & 2466 & 2205 & 2283 \\
& CoT + Python + WolframAlpha MV & 3519 & 5854 & 5096 & 5193 & 4916 \\
\cmidrule(lr){2-7}
& PAL & 328 & 442 & 426 & 394 & 398 \\
& PoT & 837 & 1140 & 1013 & 1006 & 999 \\
& ToRA & 1520 & 3801 & 2686 & 2329 & 2584 \\
& MATHSENSEI & 3915 & 5579 & 4948 & 5009 & 4863 \\
& ReAct & 1518 & 3355 & 2857 & 2496 & 2557 \\
\cmidrule(lr){2-7}
& \multitag \textbf{(Ours)} & 7952 & 18650 & 12285 & 13823 & 13178 \\
\bottomrule
\end{tabular}
}
\caption{\footnotesize Average token consumption cost (as defined in Section \ref{sec:token_consumption_cost}) per problem for all results in Table \ref{tab:main_results}}.
\label{tab:appendix_b}
\end{table*}

\subsection{\multitag API Calls Consumption}
Table \ref{tab:api_calls} reports the average number of LLM API calls per step, average number of steps used per problem, and average number of LLM API calls per problem for \multitag results in Table \ref{tab:main_results}.

\begin{table}[htbp!]
\centering
\resizebox{\textwidth}{!}{%
\begin{tabular}{l ccc|ccc|ccc|ccc}
\toprule
\multirow{2}{*}{\textbf{Model}} & \multicolumn{3}{c}{\textbf{MATH500}} & \multicolumn{3}{c}{\textbf{AIME}} & \multicolumn{3}{c}{\textbf{AMC}} & \multicolumn{3}{c}{\textbf{OlympiadBench}} \\
\cmidrule(lr){2-4} \cmidrule(lr){5-7} \cmidrule(lr){8-10} \cmidrule(lr){11-13}
& \makecell[c]{Calls \\ per \\ Step} & \makecell[c]{Steps \\ per \\ Problem} & \multicolumn{1}{c}{\makecell[c]{Calls \\ per \\ Problem}} & \makecell[c]{Calls \\ per \\ Step} & \makecell[c]{Steps \\ per \\ Problem} & \multicolumn{1}{c}{\makecell[c]{Calls \\ per \\ Problem}} & \makecell[c]{Calls \\ per \\ Step} & \makecell[c]{Steps \\ per \\ Problem} & \multicolumn{1}{c}{\makecell[c]{Calls \\ per \\ Problem}} & \makecell[c]{Calls \\ per \\ Step} & \makecell[c]{Steps \\ per \\ Problem} & \makecell[c]{Calls \\ per \\ Problem} \\
\midrule
LLaMA-3-70B & 9.99 & 2.85 & 28.49 & 13.56 & 2.79 & 37.82 & 11.86 & 2.93 & 34.72 & 12.09 & 2.96 & 35.79 \\
LLaMA-3.3-70B & 7.31 & 2.52 & 18.41 & 9.67 & 2.72 & 26.31 & 8.21 & 2.58 & 21.16 & 9.20 & 2.66 & 24.52 \\
GPT-4o & 7.98 & 2.39 & 19.11 & 11.47 & 2.69 & 30.84 & 9.44 & 2.47 & 23.33 & 10.25 & 2.56 & 26.25 \\
\bottomrule
\end{tabular}
}
\caption{\footnotesize Average number of LLM API calls per step, average number of steps per problem, and average number of LLM API calls per problem for \multitag results in Table \ref{tab:main_results}}.
\label{tab:api_calls}
\end{table}
\section{Full \multitag Trace}
\label{sec:multitag_trace}

Below is a full \multitag solution trace demonstrating \multitagg's ability to effectively leverage different tools to solve a challenging math problem. This solution trace is extracted from the GPT-4o MATH500 main results.

\textbf{Problem:}

Find the maximum value of
\[f(x,y) = x \sqrt{1 - y^2} + y \sqrt{1 - x^2},\]where \(-1 \le x,\) \(y \le 1.\)

\textbf{Step 1 (CoT):}

To find the maximum value of the function
\[f(x,y) = x \sqrt{1 - y^2} + y \sqrt{1 - x^2},\]
we can start by using the method of Lagrange multipliers. We will define a new function
\[ g(x, y) = x^2 + y^2 - 1 \]
to represent the constraint \( x^2 + y^2 = 1 \).

The method of Lagrange multipliers states that we need to solve the system of equations given by
\[ \nabla f = \lambda \nabla g \]
and the constraint equation \( g(x, y) = 0 \).

First, we will compute the gradients of \( f \) and \( g \).

The gradient of \( f \) is:
\[ \nabla f = \left( \frac{\partial f}{\partial x}, \frac{\partial f}{\partial y} \right). \]

The gradient of \( g \) is:
\[ \nabla g = \left( \frac{\partial g}{\partial x}, \frac{\partial g}{\partial y} \right). \]

Let's find these partial derivatives.

\textbf{Step 2 (Python):}

\begin{lstlisting}[frame=single, title=Python Code]
import sympy as sp

# Define the variables
x, y = sp.symbols('x y')

# Define the function f and the constraint g
f = x * sp.sqrt(1 - y**2) + y * sp.sqrt(1 - x**2)
g = x**2 + y**2 - 1

# Calculate the gradient of f
grad_f = [sp.diff(f, var) for var in (x, y)]

# Calculate the gradient of g
grad_g = [sp.diff(g, var) for var in (x, y)]

result = (grad_f, grad_g)
\end{lstlisting}

This script calculates the gradients of the functions \( f \) and \( g \). The gradients are essential for setting up the system of equations for the method of Lagrange multipliers.

\begin{lstlisting}[frame=single, title=Python Code Output]
([-x*y/sqrt(1 - x**2) + sqrt(1 - y**2), -x*y/sqrt(1 - y**2) + sqrt(1 - x**2)], [2*x, 2*y]) 
\end{lstlisting}

\textbf{Step 3 (WolframAlpha):}

\begin{lstlisting}[frame=single, title=WolframAlpha Query]
solve [-x*y/sqrt(1 - x**2) + sqrt(1 - y**2) = 2*x lambda, -x*y/sqrt(1 - y**2) + sqrt(1 - x**2) = 2*y lambda, x^2 + y^2 = 1]
\end{lstlisting}

This query solves the system of equations derived from the gradients of \( f \) and \( g \) using the method of Lagrange multipliers, along with the constraint equation \( x^2 + y^2 = 1 \).

\begin{lstlisting}[frame=single, title=WolframAlpha Result]
y = -sqrt(1 - x\^2), lambda = 0, y = sqrt(1 - x\^2), lambda = 0
\end{lstlisting}

\textbf{Step 4 (CoT):}

Given the solutions \( y = \sqrt{1 - x^2} \) and \( y = -\sqrt{1 - x^2} \) from the Lagrange multipliers method, we need to evaluate the function \( f(x, y) \) at these points to find the maximum value.

First, consider \( y = \sqrt{1 - x^2} \):
\[ f(x, \sqrt{1 - x^2}) = x \sqrt{1 - (\sqrt{1 - x^2})^2} + \sqrt{1 - x^2} \sqrt{1 - x^2} \]
\[ = x \sqrt{1 - (1 - x^2)} + (1 - x^2) \]
\[ = x \sqrt{x^2} + (1 - x^2) \]
\[ = x |x| + (1 - x^2) \]

Since \( -1 \le x \le 1 \), \( x |x| = x^2 \):
\[ f(x, \sqrt{1 - x^2}) = x^2 + (1 - x^2) \]
\[ = 1 \]

Next, consider \( y = -\sqrt{1 - x^2} \):
\[ f(x, -\sqrt{1 - x^2}) = x \sqrt{1 - (-\sqrt{1 - x^2})^2} + (-\sqrt{1 - x^2}) \sqrt{1 - x^2} \]
\[ = x \sqrt{1 - (1 - x^2)} - (1 - x^2) \]
\[ = x \sqrt{x^2} - (1 - x^2) \]
\[ = x |x| - (1 - x^2) \]

Again, since \( -1 \le x \le 1 \), \( x |x| = x^2 \):
\[ f(x, -\sqrt{1 - x^2}) = x^2 - (1 - x^2) \]
\[ = x^2 - 1 + x^2 \]
\[ = 2x^2 - 1 \]

The maximum value of \( 2x^2 - 1 \) over the interval \( -1 \le x \le 1 \) is 1, which occurs at \( x = \pm 1 \).

Thus, the maximum value of \( f(x, y) \) is \(\boxed{1}\).

\end{document}